\documentclass{article}      

\usepackage{graphicx}
\usepackage{times}
\usepackage{amssymb}
\usepackage{amsmath}
\reversemarginpar
\usepackage{geometry}
\usepackage{fancyhdr}
\fancypagestyle{firststyle}
{
   \fancyhf{}
   \fancyhead[C]{This paper has been published in Journal of Signal Processing Systems. Please cite it as:\\ \textbf{Amirhossein Tavanaei and Anthony S. Maida, "Training a Hidden Markov Model with a Bayesian Spiking Neural Network", Journal of Signal Processing Systems, (2016), 1-10.}\\
   The final publication is available at Springer via: 
   \textbf{http://dx.doi.org/10.1007/s11265-016-1153-2}}
}
\begin{document}

\title{Training a Hidden Markov Model with a Bayesian Spiking Neural Network}

\author{Amirhossein Tavanaei and Anthony S. Maida\\
\textit{\small{The Center for Advanced Computer Studies}}\\
\textit{\small{University of Louisiana at Lafayette, LA 70504, USA}}\\
\textit{\small{tavanaei@louisiana.edu, maida@cacs.louisiana.edu}}}

\date{}
\maketitle

\thispagestyle{firststyle}

\begin{abstract}
It is of some interest to understand how statistically based mechanisms for signal
processing might be integrated with biologically motivated mechanisms such as neural networks.
This paper explores a novel hybrid approach for classifying segments of sequential data,
such as individual spoken works.
The approach combines a hidden Markov model (HMM) with a spiking neural network (SNN).
The HMM, consisting of states and transitions, forms a fixed backbone with nonadaptive
transition probabilities.
The SNN, however, implements a biologically based Bayesian computation that derives from the spike timing-dependent
plasticity (STDP) learning rule.
The emission (observation) probabilities of the HMM are represented in the SNN 
and trained with the STDP rule.
A separate SNN, each with the same architecture, is associated with each of the states
of the HMM\@.
Because of the STDP training, each SNN implements an expectation maximization algorithm to
learn the emission probabilities for one HMM state.

The model was studied on synthesized spike-train data and also on spoken word data.
Preliminary results suggest
its performance compares favorably with other biologically motivated approaches.
Because of the model's uniqueness and initial promise, it warrants further study.
It provides some new ideas on how the brain might implement the equivalent of an HMM
in a neural circuit.

\end{abstract}

\section{Introduction}
\label{intro}


In some settings, it is desirable to have a biologically motivated approach for classifying
segments of sequential data, such as spoken words.
This paper examines a novel hybrid approach towards such data classification.
The approach uses two components.
The first is the hidden
Markov model (HMM)~\cite{Bishop2006a} and the second is a biologically
motivated spiking neural network (SNN)~\cite{Maass1997a,Samanwoy2009a} that approximates expectation maximization learning (EM)
\cite{Nessler2009a,Nessler2013a}.
In addition to the intrinsic interest of exploring statistically based biologically motivated approaches to machine learning,
the approach is also attractive because of its possible realization on special purpose hardware
for brain simulation \cite{Modha2014a} as well as fleshing out the details of a large-scale
model of the brain \cite{Eliasmith2012a}.

HMMs are widely used for sequential data classification tasks, such as speech recognition \cite{Rabiner1989b}.
There have been earlier efforts to build hybrid HMM/neural network models \cite{Bourlard1988a,Niles1990a,Bengio1992a}.
In this work, the hybrid approach was motivated by the insight that ANNs perform well for non-temporal classification
and approximation while HMMs are suitable for modeling the temporal structure of the speech signal.
More recent work has used more powerful networks, such as deep belief networks and 
deep convolutional networks, for acoustic modeling of the 
speech signal \cite{Hinton2011a,Abdel-Hamid2012a,Abdel-Hamid2013a,Sainath2013a}.
While these efforts have met with considerable practical success, they are not obviously biologically motivated.
In part, our work differs from the previous work in that we use a biologically motivated
SNN\@.

Formally, an HMM consists of a set of discrete states, a state transition probability matrix,
and a set of emission (observation) probabilities associated with each state.
The set of trainable parameters in an HMM can be the initial state probabilities, the transition
probabilities, and the emission probabilities.
This paper limits itself to training the emission probabilities using an SNN\@.
This is consistent the approaches of the above-mentioned earlier work.

Our work is directly influenced by the 
important prior work on how an HMM might be implemented in a cortical microcircuit
was performed by \cite{Kappel2014a}.
The cortical microcircuit is a repeated anatomical motif in the neocortex who some
have argued is the next functional level of description above the single neuron \cite{Mountcastle1997a}.
In its most simplified form, the microcolumn can be modeled as a recurrent neural network
with lateral inhibition.
Kappel et al. \cite{Kappel2014a} have recently shown that,
with appropriate learning assumptions,
a trainable HMM can be realized within this microcircuit.
The contribution of the present work is to unwrap this microcircuit into a more discernable
HMM\@.
The motivation for our approach is to recognize the fact that there are many
ways to potentially realize an HMM in the brain and we seek a model that may
be developed in future work but that does not burn any bridges or make unnecessary commitments.

The motivation for this study is not so much to build the highest performing HMM-based
classifier as it is to imagine how: 1) an HMM might be realized in the brain, and 2) be implemented
in brain-like hardware.

\section{Background}
\label{sec:1}

Since this research combines spike timing-dependent plasticity
(STDP) learning with HMM classifiers, the next subsections provide
background on each topic.

\subsection{Spike timing-dependent plasticity}
The phenomenon of STDP learning in the brain has been known for at least two decades \cite{Markram2011a}.
STDP modifies the connection strengths between neurons at their contact points (synapses).
Spikes travel from the presynaptic neuron to the postsynaptic neuron via synapses.
The strength of the synapse, represented by a scalar weight, modulates the likelihood of a
presynaptic spike event causing a postsynaptic event.
The weight of a synapse can be modified (plasticity) by using learning rules that incorporating information
locally available at the synapse (for example, STDP).

Generically, an STDP learning rule operates as follows.
If the presynaptic neuron fires briefly before the postsynaptic
neuron, then the synaptic weight is strengthened.
If the opposite happens, the synaptic weight is weakened.
Such phenomena have been experimentally observed in many brain areas \cite{Dan2006a,Corporale2008a}.
A simple intuitive interpretation of this empirically observed constraint is that the synaptic strength is increased
when the presynaptic neuron could have played a causal role in the firing of the postsynaptic
neuron. The strength is weakened if causality is violated.

Probabilistic interpretations of STDP that could form a theoretical link to machine
learning have emerged in the past decade.
Most relevant to this paper are the following.
Nessler et al. \cite{Nessler2009a,Nessler2013a} developed a version of STDP to compute
EM within a spiking neural circuit.
Building on this, Kappel et al. \cite{Kappel2014a} built an HMM within a recurrent SNN\@.
The recurrent SNN coded for all of the states in the HMM as well as implementing the 
learning.
This was a significant hypothesis from a brain-simulation because of its
very strong claim, that a cortical microcircuit may implement a full-blown HMM\@.
The hypothesis is also
highly, perhaps overly, committed from an engineering perspective.

The present approach seeks to use the right tool for the right job while still
linking it to a biomorphic framework.
Specifically, we encode states using an HMM but associate a separate copy of the modified version of a
trainable Nessler-type SNN with each state.
The purpose of the SNN is to learn the emission (observation) probabilities for that state.
In future work, one may find other effective ways to fully encode an HMM model as an SNN,
but it may not necessarily be the approach taken in \cite{Kappel2014a}.

\subsection{Hidden Markov model}
\label{subsec:HMM}
\begin{figure}
\begin{center}
  \includegraphics[viewport=1.0in 8.3in 5in 10in,clip=true,scale=0.8]{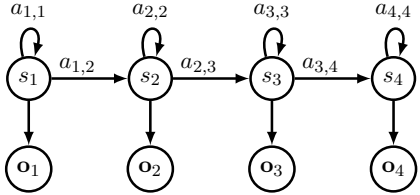}
\end{center}
\caption{A four-state, left-to-right HMM, labeled with fixed transition
probabilities $a_{ij}$, where the state index is
allowed to increase at most by one for each transition.
The number of states $P$ is four.}
\label{fig:HMMDiagram}
\end{figure}

Successive observations of sequential data, such as occurs in speech spectrograms,
are highly correlated.
The correlation often drops significantly between observations that are sequentially distant.
An effective way to classify sequential data is to use a markov chain of latent variables (states),
otherwise known as an HMM\@.
The HMM describes the data as a first-order Markov chain that assumes the probability
of the next state is independent of all of the previous states, given the current
state.
Fig.~\ref{fig:HMMDiagram} shows a four-state, left-to-right HMM, whose initial
state is \textsf{s1}.
In the figure, arrows entering nodes that are labeled ``\textsf{s}'' are state-transition
probabilities and arrows entering nodes labeled ``\textsf{o}'' represent the causal relationship
between a state and an observation.

The HMM allows calculation of the probability of a given observation sequence 
of feature vectors
$\mathbf{O} = [\mathbf{o}_1 \ldots \mathbf{o}_M]$,
in a structure consisting of states $\mathbf{S} = \{s_1 \ldots s_P\}$,
initial state probabilites $\mathbf{\Gamma}$,
state transition probabilities $\mathbf{A}$,
and emission probabilities $\mathbf{B}$.
The probability of a particular observation sequence is given by
\begin{equation}
\mathit{Pr}(\mathbf{O} | \boldsymbol\lambda) = \prod_{m=1}^{M} \mathit{Pr}(\mathbf{o}_m | s_m, \boldsymbol\lambda) ,
\label{eqn:mainHMM}
\end{equation}

\noindent
where $\boldsymbol\lambda$ denotes the set of model parameters and $s_m$ denotes the state of the HMM when 
the observation occurs.
The emission probabilities are given by $\mathit{Pr}(\mathbf{o}|s, \boldsymbol\lambda)$ and these are the
parameters learned by the SNN\@.
We will have occasion to use the symbols $s_m$ versus $s_p$.  
The former means the state of the HMM when observation $m$ occurs.
The latter simply means state $p$ of the HMM. 

To train and adjust the model parameters in an HMM, the expectation maximization (EM) approach (also known as the Baum-Welch algorithm in the HMM) is used \cite{Rabiner1993a}. In this paper we assumed fixed transition probabilities, $\mathbf{A}$, for all of the states. 

\section{Probability Computation}
\subsection{Bayesian computation}

In a Bayesian framework,
a posterior probability distribution is obtained by multiplying the prior
probability with the likelihood of the observation and renormalizing.
Recent studies have shown that the prior and likelihood models of observations
can be represented by appropriately designed neural networks \cite{Deneve2008a}.
One such network is a spiking winner-take-all (WTA) network\@.
Nessler et al. \cite{Nessler2009a,Nessler2013a} showed that a version of the STDP rule
embedded in an appropriate SNN can perform Bayesian computations.
\subsection{Gaussian mixture model}

The most general representation of the probability distribution function in the HMM state is a finite mixture of the Gaussian distributions (GMM) with mean vector, $\boldsymbol{\mu}$, covariance matrix, $\boldsymbol{\Sigma}$, and mixture coefficients, $\boldsymbol{\pi}$. Each HMM state has its own mixture distribution.
The probability of observation $\mathbf{o}_m$
occurring in HMM state $s_p$ is given by

\begin{equation}
\mathit{Pr}(\mathbf{o}_m | s_p) = \sum_{k=1}^K \pi_{k} \cdot \mathcal{N}(\mathbf{o}_m|\boldsymbol{\mu}_k,\mathbf{\Sigma}_k), \;\;\;\; 1\le p\le P
\end{equation}
where $P$ is the number of HMM states, $\pi_{k}$ is the mixing parameter, and $K$ is the number of distributions
in the mixture.

In our model,
the emission distributions for the HMM states approximately implement the Gaussian mixture distributions.
There is a separate mixture distribution associated with each state, corresponding
to a separate SNN\@.
The SNN learns distribution
parameters via STDP\@.

\begin{figure}
\begin{center}
  \includegraphics[viewport=1.0in 4.2in 7.5in 7in,clip=true,scale=0.6]{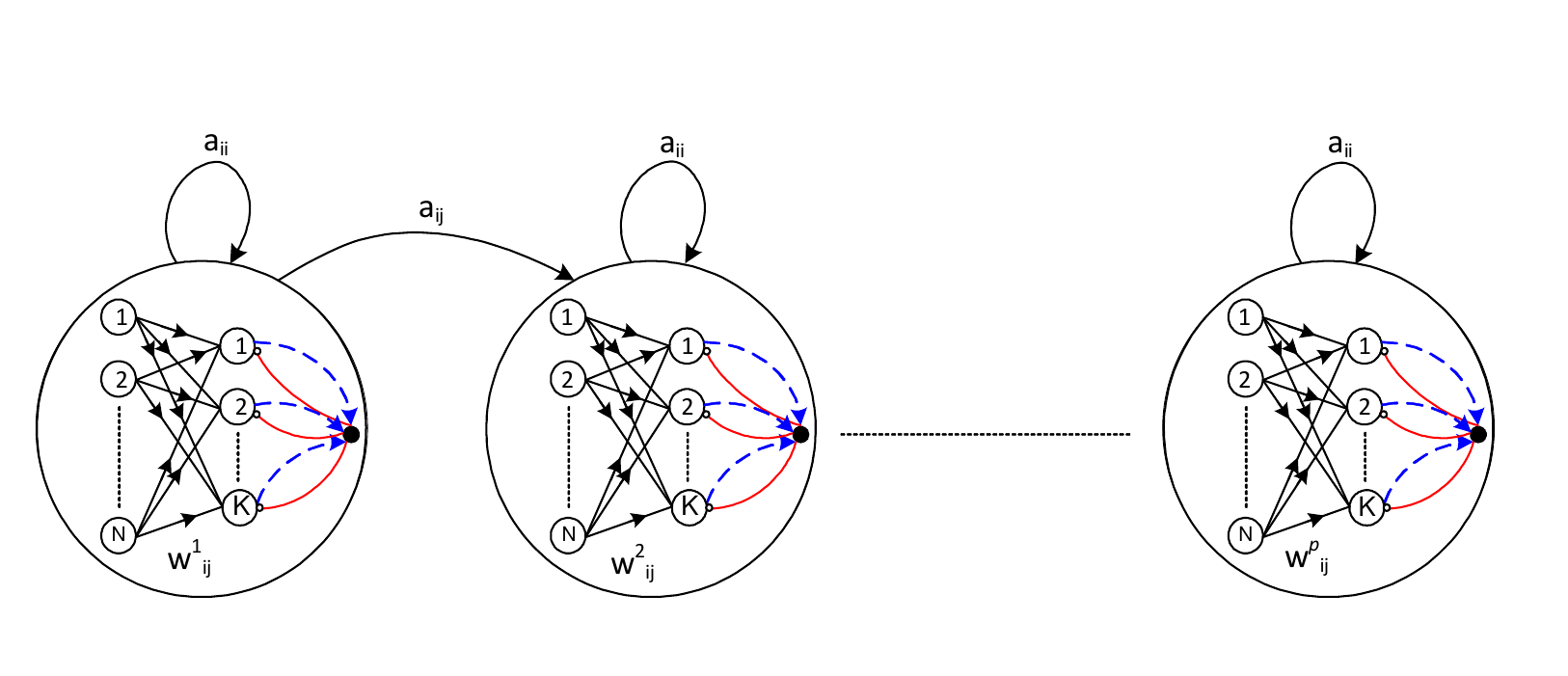}
\end{center}
\caption{The hybrid model containing $P$ HMM states.
Each HMM state has an associated SNN. Each SNN has $N$ input units, $y$, and $K$ output units, $z$, and a global inhibition (black circle). The output units receive $N$ (as number of the feature values of an observation) spike trains from input neurons. The parameters $a_{ij}$ between states are transition probabilities of the HMM. 
}
\label{fig:stateTransDiagram}
\end{figure}

\section{Training Method}

The SNN trains the parameters for the emission distributions and each HMM state
has a separate SNN as shown in Fig.~\ref{fig:stateTransDiagram}. 
Fig.~\ref{fig:SNNarchitecture} shows the SNN architecture in detail.
Additionally, Nessler et al. \cite{Nessler2013a} showed their STDP learning approximates EM. 
Therefore, the proposed SNN architecture is able to implement the GMM in each state (described in section 4.2).

\subsection{Network architecture}
The SNN has two layers of stochastic units (neurons) that generate Poisson spike trains.
The $y$ units in the first layer encode input feature vectors to be classified.
The second layer is composed of $z$ units that represent classification categories after the network is trained.
The layers are fully feedforward connected from layer $y$ to $z$ by weights trained according to the
STDP rule given in the next subsection.
Besides the feedforward connections,
the $z$ units obey a winner-take-all discipline implemented
by a global inhibition signal initiated by any of the $z$ units. 

\begin{figure}
\begin{center}
  \includegraphics[viewport=1.0in 6.3in 6in 9.5in,clip=true,scale=0.6]{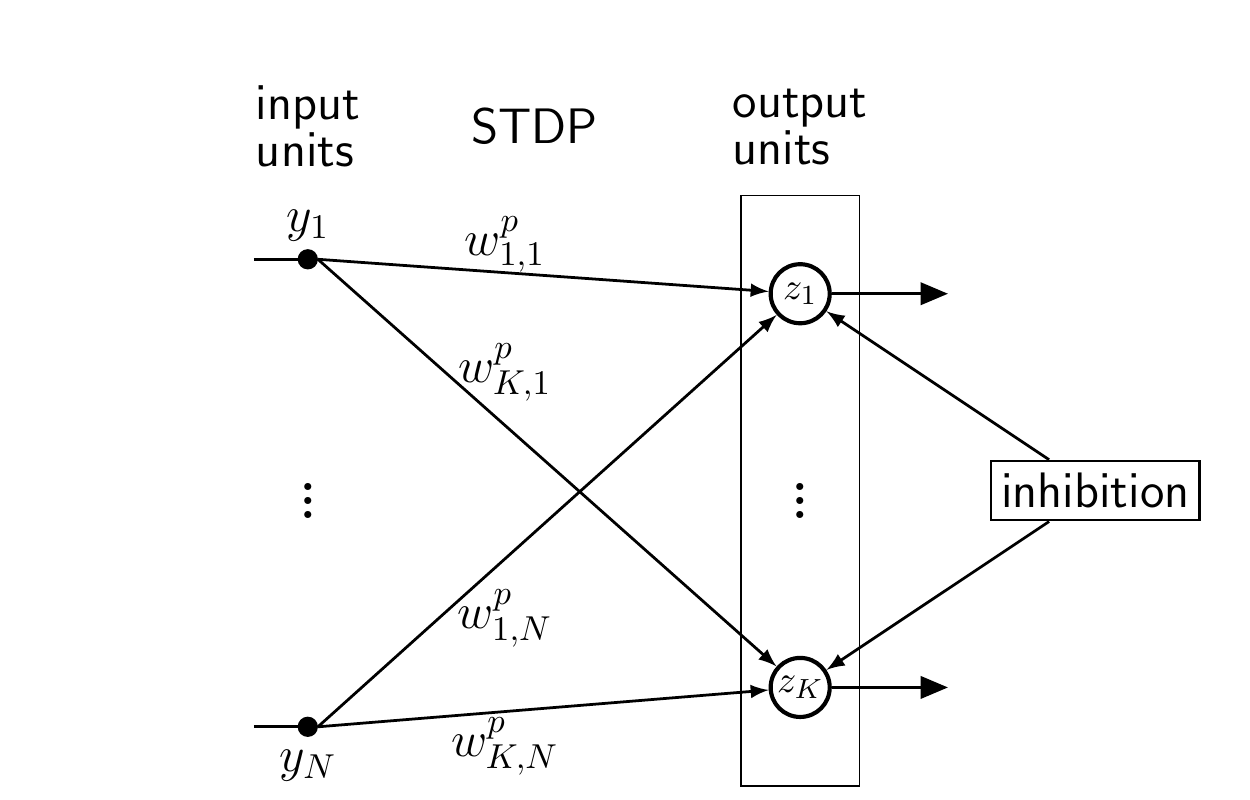}
\end{center}
\caption{SNN architecture to compute emission probabilities.
The architecture is identical for each state but the weight matrix is state specific.}
\label{fig:SNNarchitecture}
\end{figure}

The number of $y$ units in Fig.~\ref{fig:SNNarchitecture}, $N$, shows the feature vector dimension. The $z$ units specify the output neurons detecting the samples in $K$ different clusters. 
The number of output neurons, $K$, manipulates the model flexibility in controlling the signal variety in one segment (analogous to the number of distributions in a GMM). The number of states, $P$, determines the number of segments in a sequential signal. For example, in spoken word recognition, it can be considered as the number of phoneme bigrams.

The spiking activity of a unit in the $z$ layer is governed by an inhomogeneous Poisson process.
The rate parameter for this process is controlled by the postsynaptic potential (PSP) input to the $z$ unit.
The PSP represents the sum of the synaptic effects coming into the $z$ unit.
The instantaneous firing rate of unit $k$
is given by $r_k(t)$ 
and is defined below
\begin{equation}
r_k(t) = \exp( {\mathit{psp}_k(t)} ).
\label{eq:exp_rk}
\end{equation}

The $\mathit{psp}$ itself is the sum of 
the excitatory inputs into $k$ from the $y$ layer and a global inhibitory input.
These are denoted respectively as $u_k(t)$ and $I(t)$.
Thus, the $\mathit{psp}$ for unit $k$ at time $t$ is
\begin{equation}
\mathit{psp}_k(t) = u_k(t) + I(t).
\label{eq:psp_components}
\end{equation}

\noindent
$u_k$ encodes the composite stimulus input signal to unit $k$.
The input signal is provided by the
$y$ units, which  encode the input feature vector.
The quantity $u_k$ is a linear weighted sum of the excitatory postsynaptic potentials (EPSPs)
provided by the $y$ units as shown below
\begin{equation}
u_k(t) = w_{k0}(t) + \sum_{i=1}^N \mathit{epsp}_i^k(t) \cdot w_{ki}(t).
\label{eq:exp_uk}
\end{equation}

\noindent
There are $N$ `excitatory' units in the $y$ layer.
$\mathit{epsp}_i^k$ denotes the component of the EPSP
of $k$ that originates with unit $i$ in the $y$ layer.
$w_{k0}$ is the bias weight to unit $k$.
The $w_{ki}$ are the weights from units in the $y$ layer to unit $k$ in the $z$ layer.
The weights and bias are time dependent because their values can change while the
network is learning.
$\mathit{epsp}_i^k$ is defined by
\begin{equation}
\mathit{epsp}_i^k(t) = 
    \left\{
    \begin{array}{ll}
    1&\mathrm{if}\; i\;\mathrm{fired\;during\;interval} \;\left[t-\sigma,t\right]\\
    0&\mathrm{otherwise}.
    \end{array}
    \right.
\label{eq:exp_epsp}
\end{equation}
\noindent
The quantity $\mathit{epsp}_i^k(t)$
has a value of 1 at time $t$ if and only if unit $i$ has fired in the previous $\sigma$ milliseconds.
In the simulations, $\sigma=5$.

\subsection{Training}
\subsubsection{GMM learning approximation by SNN}
\paragraph{Posterior probability}

The Gaussian distribution over the dataset, $\mathbf{y}$, is defined as

\begin{equation}
\mathcal{N}(\mathbf{y}|\boldsymbol{\mu},\boldsymbol{\Sigma})=\frac{1}{(2\pi )^{N/2} |\boldsymbol\Sigma|^{1/2}}e^{-0.5 (\mathbf{y}-\boldsymbol\mu)^T\boldsymbol\Sigma^{-1}(\mathbf{y}-\boldsymbol\mu)}
\label{eq:normal}
\end{equation}
where $\boldsymbol\Sigma$ and $\boldsymbol\mu$ are covariance matrix and mean vector respectively.   
In the Gaussian mixture model with $K$ mixtures $z_1 \ldots z_K$, $z_k \in \{0, 1\}$, $\sum z_k = 1$,
the probability of a sample, $\mathbf{y}_r$, is derived as follows:
\begin{equation}
\mathit{Pr}(\mathbf{y}_r|z_k=1)=\mathcal{N}(\mathbf{y}_r|\boldsymbol\mu_k,\boldsymbol\Sigma_k),
\label{eq:p_y}
\end{equation}
\begin{equation}
\mathit{Pr}(\mathbf{y}_r|\mathbf{z})=\prod_{k=1}^K \mathcal{N}(\mathbf{y}_r|\boldsymbol\mu_k,\boldsymbol\Sigma_k)^{z_k},
\label{eq:p_ys}
\end{equation}
\begin{equation}
\mathit{Pr}(\mathbf{y}_r)=\sum_\mathbf{z} \mathit{Pr}(\mathbf{z})\mathit{Pr}(\mathbf{y}_r|\mathbf{z})=\sum_{k=1}^K \pi_k \mathcal{N}(\mathbf{y}_r|\boldsymbol\mu_k,\boldsymbol\Sigma_k),
\label{eq:gmm}
\end{equation}
\begin{equation*}
\mathrm{where} \ \ \sum_k \pi_k=1, \ \ \ 0\leq \pi_k \leq 1.
\label{eq:gmm_constraint}
\end{equation*}
From Eq.~\ref{eq:p_y} and Eq.~\ref{eq:gmm} we have
\begin{equation}
\mathit{Pr}(z_k=1|\mathbf{y}_r)=R(z_{kr})= \frac{\pi_k \mathcal{N}(\mathbf{y}_r|\boldsymbol\mu_k,\boldsymbol\Sigma_k)}{\sum_{j=1}^K \pi_j \mathcal{N}(\mathbf{y}_r|\boldsymbol\mu_j,\boldsymbol\Sigma_j)}.
\label{eq:R}
\end{equation}
The conditional probability $\mathit{Pr}(z_k | \mathbf{y}_r)$ can also be written as $R(z_{kr})$ which represents the responsibility~\cite{Bishop2006a}. 
To simplify the equations, assume that the samples are independent from each other in which $\mathbf{\Sigma}=\textbf{I}$. So,
\begin{equation}
\mathit{Pr}(z_k=1|\mathbf{y}_r)=R(z_{kr})=\frac{\pi_k B e^{-0.5(\mathbf{y}_r-\boldsymbol\mu_k)^T(\mathbf{y}_r-\boldsymbol\mu_k)}}{\sum_{j=1}^K \pi_j B e^{-0.5(\mathbf{y}_r-\boldsymbol\mu_j)^T(\mathbf{y}_r-\boldsymbol\mu_j)}}.
\label{eq:R_simple}
\end{equation}
The equation above reaches its maximum when $\mathbf{y}_r=\boldsymbol\mu_k$. Now, if we suppose the synaptic weight vector of neuron $k$, $w_{ki},$ ($i=1 \ldots N$) is an $N$ dimensional vector which approximates the $\boldsymbol\mu_k$, the similarity between the sample and mean (negation of distance measure), $-0.5(\mathbf{y}-\boldsymbol\mu)^T(\mathbf{y}-\boldsymbol\mu)$, can be replaced by similarity measure between $\textbf{y}$ and $\textbf{w}$ as $\textbf{w}^T\textbf{y}$ (projection of the sample vector on the weight vector reaches the maximum when they are in a same direction). Thus,
\begin{equation}
R(z_{kr})=C\frac{\pi_k e^{\mathbf{w}_k^T \mathbf{y}_r}}{\sum_{j=1}^K \pi_j e^{\mathbf{w}_j^T\mathbf{y}_r}}
\label{eq:R_snn}
\end{equation}
where $C$ is a constant. $\pi_k$ and $K$ denote the mixture coefficients and number of the mixture distributions, respectively.

\paragraph{STDP rule specification}
In the training process of the GMM using EM, we have
\begin{equation}
\boldsymbol\mu_k^{new}=\frac{\sum_{s=1}^M R(z_{ks})\mathbf{y}_s}{\sum_{s=1}^M R(z_{ks})},
\label{eq:em}
\end{equation}
\begin{equation}
\pi_k^{new}=\frac{\sum_{s=1}^M R(z_{ks})}{M}
\label{eq:em_pi}
\end{equation}
where $M$ is the number of training samples. 
In the proposed SNN, $k$ is the output neuron that has just fired.
$\mathbf{w}_k$, which is supposed as $\boldsymbol\mu_k$ in the GMM and already represents the previous samples in this cluster, should be updated based on the new samples. 
Instead of calculating the average value of the samples (Eq.~\ref{eq:em}), new synaptic weight, $\mathbf{w}_k^{new}$ (or $\boldsymbol\mu_k^{new}$), is updated by $\mathbf{w}_k+f(\mathbf{y}_r)$ where $f(\mathbf{y}_r)$ has $N$ positive and negative numbers corresponding to $\mathbf{y}_r(i)=1$ and $\mathbf{y}_r(i)=0$ respectively. 
Thus, the new $\mathbf{w}_k$ is updated using $R(z_{kr})$ (which causes a neuron to fire) and input presynaptic spikes. 
For this purpose we use a modified version of the Nessler's (2013) STDP learning rule. STDP is an unsupervised learning rule.
Following \cite{Nessler2013a},
weight adjustments occur
exactly when some $z$ unit $k$ emits a spike.
When a unit $k$ fires, the incoming weights to that unit are subject
to learning according to the STDP rule given in Eq.~\ref{eq:NesslerSTDP_wts}.
For each weight, one of two weight-change events occurs, either LTP (strengthening) or LTD (weakening).
The weight values are constrainted to be in the range [-1 1].
\begin{equation}
\Delta w_{ki} = 
     \left\{
     \begin{array}{ll}
     e^{-w_{ki}+1}-1 & \mathrm{if}\;\mathit{epsp}_i(t^\mathrm{f})=1\\
     -1           & \mathrm{otherwise}.
     \end{array}
     \right.
\label{eq:NesslerSTDP_wts}
\end{equation}

\noindent
The first case above describes LTP (positive) and the second case describes LTD (always $-1$)\@.

Another parameter of the GMM is the mixture coefficient $\pi_k$ which is obtained by Eq.~\ref{eq:em_pi}. The bias weight of the proposed SNN, $w_{k0}$, represents average firing of the neuron $k$ over data occurrences. Therefore, if the neuron fires, its bias weight increases, otherwise it decreases. For this purpose we use a modified version of the Nessler's (2013) STDP learning rule analogous to Eq.~\ref{eq:NesslerSTDP_wts} as follows:
\begin{equation}
\Delta w_{k0}=z_ke^{-w_{k0}+1}-1.
\label{eq:w0}
\end{equation}

\noindent
$\Delta w_{ki}$ denotes a weight adjustment that is modulated by another
rate parameter $\eta_k$.
Specifically,

\begin{equation}
w_{ki}^\mathrm{new} = w_{ki} + \eta_k \Delta w_{ki}.
\label{eq:wNewFormula}
\end{equation}

\noindent
That is,
there is a rate parameter $\eta_k$ for each $z$ unit with $N_k$ as the number of times the unit has fired,
starting with 1.
It satisfies the constraint
\begin{equation}
\eta_k \propto \frac{1}{N_k}.
\label{eq:etaKformula}
\end{equation} 
 
By considering the mixture coefficient in the SNN, $\pi_k^{\mathrm{snn}}$, to be defined as follows:
\begin{equation}
\pi_k^{\mathrm{snn}}=\frac{e^{w_{k0}}}{D},
\label{eq:pi_snn}
\end{equation}
\begin{equation}
D=\sum_j e^{w_{j0}}, \nonumber
\label{eq:pi_snn_norm}
\end{equation}
the constraints on the mixture coefficients in Eq.~\ref{eq:gmm} are fulfilled. Finally, by combining Eq.~\ref{eq:R_snn} and Eq.~\ref{eq:pi_snn} we have
\begin{equation}
R(z_{kr})=A\frac{\frac{e^{w_{k0}}}{D} e^{\mathbf{w}_k^T\mathbf{y}_r}}{\sum_{j=1}^K \frac{e^{w_{j0}}}{D} e^{\mathbf{w}_j^T\mathbf{y}_r}},
\label{eq:final1}
\end{equation}
\begin{equation}
\mathit{Pr}_k(z \ \mathrm{fires}|\mathbf{y}_r)=R(z_{kr})=A\frac{e^{\mathbf{w}_k^T\mathbf{y}_r+w_{k0}}}{\sum_{j=1}^K e^{\mathbf{w}_j^T\mathbf{y}_r+w_{j0}}}.
\label{eq:finalgmmsnn}
\end{equation}

\paragraph{Training procedure}
Since there is a separate SNN for each state,
the observation function $Pr(\mathbf{o}_m | s_m, \boldsymbol\lambda)$ can be trained separately for each state.
The training procedure begins with a set of feature vectors and initial weights.
Randomly selected feature vectors from the sample to be recognized are
presented to the network for some number of training trials.

To the extent that the feature vectors are similar to previous observations,
a subset of output neurons fire and the synaptic weights are updated according to
Eq.~\ref{eq:NesslerSTDP_wts} through Eq.~\ref{eq:wNewFormula}.
A new feature vector, which is different from previous vectors, stimulates a new set 
of output neurons to fire.
This strategy imposes an unsupervised learning method within the SNN to categorize
the data in one state.

\paragraph{Extracting a probability value from the SNN}
We will let $Pr_{\mathrm{snn}_p}(t)$ denote the probability that the SNN input 
at simulation step $t$ is
of the category that the network recognizes.
This value is the maximum of the output units after normalization as described below (simplified representation of Eq.~\ref{eq:finalgmmsnn})
\begin{equation}
Pr_{\mathrm{snn}_p}(t) = \max_{k\in K} \frac{e^{u_k(t)}}{Z},
\label{eq:SNN_output_prob}
\end{equation}

\noindent
where $Z$ is a normalizer.

\section{Experiments and Results}
\subsection{Synthesized spatio-temporal spike patterns}

This experiment modeled a pattern classification task that
used four spatio-temporal spike sub-patterns to build a larger pattern.
Each sub-pattern consisted of 80 neurons that simultaneously emitted Poisson spike trains.
The duration of all spike trains within a sub-pattern was $T=20$ ms.
This was called a spatio-temporal pattern because the 80 neurons compose the
spatial dimension~\cite{Dayan2001a}.
Target patterns were obtained by concatenating the four sub-patterns.

\begin{figure}
\begin{center}
  \includegraphics[viewport=1.8in 4.2in 7.5in 7in,clip=true,scale=1.0]{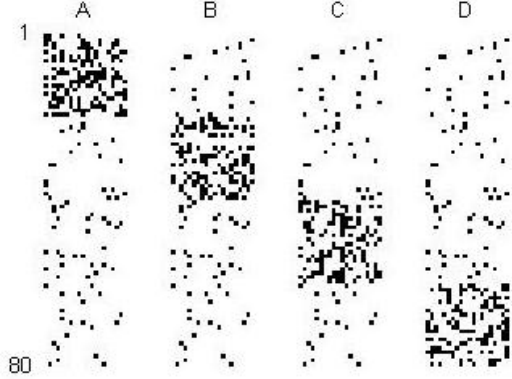}
\end{center}
\caption{One example of each of the four input sub-patterns.
Each sub-pattern consists of 80 Poisson spike trains whose duration is 20 ms.
Background firing rate is 50 Hz.
Firing rate for neurons carrying information is 340 Hz.}
\label{fig:spikeTrainsExp1}
\end{figure}

Examples of the sub-patterns denoted A, B, C, and D are shown in Fig.~\ref{fig:spikeTrainsExp1}.
Different instances of a specific sub-pattern, such as A, will have different spike trains because
of the Poisson sampling.
Each row shows a spike train for a single neuron.
For each sub-pattern, twenty of the neurons fire at 340 Hz and the remaining sixty neurons
fire at 50 Hz.
The high frequency spike trains were deemed information-containing and the low-frequency
spike trains were considered background noise.

\begin{figure}
\begin{center}
  \includegraphics[viewport=0.0in 0.0in 7.5in 5in,clip=true,scale=0.5]{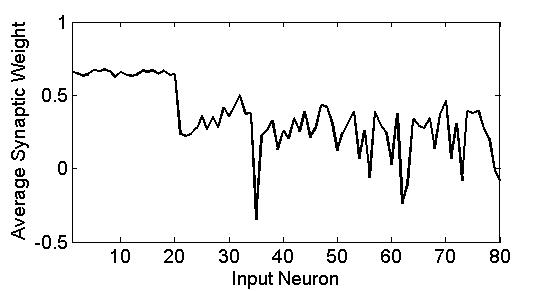}
\end{center}
\caption{Weights for sub-pattern A after training. The results for the other sub-patterns were analogous.}
\label{fig:trainedSubpatternWeights}
\end{figure}

\paragraph{Training phase}
Only sub-patterns were trained.
The SNN training was unsupervised according to STDP explained earlier.
An SNN had 80 input units corresponding to each of the 80 spike trains forming the pattern.
The network had eight output units
to allow within category diversity.
For each sub-pattern, one SNN was trained for ten iterations using
STDP\@.
One iteration meant that the network was allowed to run for $T=20$ ms with STDP enabled
and the input neurons maintained Poisson firing rates according to their location within the sub-pattern.
Synaptic weights were randomly initialized before training.
Fig.~\ref{fig:trainedSubpatternWeights} shows average the synaptic weights after training
for sub-pattern A\@.
The weights for high-firing-rate spike trains 1--20 are clearly distinguishable
from the weights for low-firing-rate spike trains 21--80.
The results for training the other sub-patterns were analogous.
The plot shows that information can be detected in the presence of noise.
The average synaptic weights, $\textbf{w}_\mathrm{state}$, were calculated by averaging over the eight output units which is shown in Eq.~\ref{eq:AvgW}.

\begin{equation}
\mathbf{w}_\mathrm{state}=\frac{1}{K}\sum_{k=1}^K \mathbf{w}_k \cdot w_{k0}.
\label{eq:AvgW}
\end{equation}

\noindent
Recall that the bias learns to represent the average firing rate.

Patterns to be classified were built from a sequence of the four sub-patterns. 
Each sub-pattern in a sequence corresponded to one HMM state.
A collection of four HMMs, each with four states corresponding to the pattern length,
were used to recognize four target patterns ABCD, DCBA, ABDC, and BACD.

Table~\ref{tab:artificialDataResults} shows the performance results on this data set.
In the table, \textit{desired} means the pattern that was presented and
\textit{recognized} means the HMM with the highest probability output. 
The probabilities along the diagonal (correctly classified) are much higher than the other probabilities in each column. 
Therefore, the proposed model shows initial promising results in categorizing a simple set of the synthesized spatio-temporal patterns.

\begin{table}
\centering
\caption{Classification probabilities for the artificial patterns: \textsf{ABCD},
\textsf{DCBA}, \textsf{ABDC}, and \textsf{BACD}.
}
\label{tab:artificialDataResults}       
\begin{tabular}{lrrrr}
\hline\noalign{\smallskip}
Recognized/desired & \textsf{ABCD} & \textsf{DCBA} & \textsf{ABDC} & \textsf{BACD}\\
\noalign{\smallskip}\hline\noalign{\smallskip}
\textsf{ABCD} & \textbf{0.442} & 0.112 & 0.227 & 0.219\\
\textsf{DCBA} & 0.145 & \textbf{0.572} & 0.140 & 0.143\\
\textsf{ABDC} & 0.254 & 0.124 & \textbf{0.495} & 0.126\\
\textsf{BACD} & 0.249 & 0.123 & 0.128 & \textbf{0.500}\\
\noalign{\smallskip}\hline
\end{tabular}
\end{table}

\paragraph{Details of the recognition mechanism}
During the recognition phase, 
we used a set of four $P$-state HMMs ($P=4$) for each of the target patterns.
The four trained SSNs are associated with the appropriate HMM state.
The HMM with highest probability for a given input
sequence was taken as the best match to the input signal.
For each HMM, the probability of an observation seuquence, $\mathbf{O}$, was calculated 
by expanding Eq.~\ref{eqn:mainHMM} as follows:
\begin{subequations}
\label{eq:HMMrecognitionEqns}
\begin{align}
\mathit{Pr}(\mathbf{O} | \boldsymbol\lambda) &= Pr(s_m =1) \prod_{p=2}^4 a_{p-1,p} \cdot Pr( s_m = p ) \\ &
\mathit{Pr}( s_m = p) = \prod_{t=1}^{T=20}\mathit{Pr}_{\mathrm{snn}_p}(t).
\end{align}
\end{subequations}

\noindent
$\mathit{Pr}_{\mathrm{snn}_p}(t)$ is defined in Eq.~\ref{eq:SNN_output_prob}.
The $a_{p-1,p}$'s and $a_{p,p}$'s are all set to $0.5$ (fixed transitions).
All other $a_{ij}$'s are set to zero.
In this experiment, total pattern duration was 80 ms corresponding to a concatenated sequence
of four sub-patterns ($T=20$ ms). 
The state probability, $\mathit{Pr}( s_m = p)$, in Eq.~\ref{eq:HMMrecognitionEqns} is obtained by multiplying the particular state probabilities in $T=20$ sequential time steps (1 ms separation). 
$T$ is the duration for the Poisson spike trains. 
From Eq.~\ref{eq:exp_uk} and Eq.~\ref{eq:SNN_output_prob}, the state probability should have an exponential form as
\begin{equation}
\mathit{Pr}( s_m = p)= e^{\textbf{w}_s^T \cdot \sum_{t=1}^{T=20}\textbf{y}(t)}
\label{eq:provePoisson}
\end{equation}   
where $\textbf{w}_s$ is the selected weight vector with maximum probability value in Eq.~\ref{eq:SNN_output_prob}. 
$\sum_{t=1}^{T=20}\textbf{y}(t)$ reports the \textit{Poisson process rate} $\times\; T$ which can be interpreted as the feature values of an observation. 
Therefore, it reversely shows the statistical similarity between two numerical vectors $\textbf{w}_s$ and $\textbf{y}$ discussed in section (4.2.1, \textit{posterior probability}). 

The algorithm for training the hybrid HMM/SNN model and classifying the sequential patterns is shown in Fig.~\ref{fig:modelAlgorithm}.
For this example, Lines~2 and 3 were not needed because the sub-patterns were already extracted.
%

\begin{figure}
\begin{center}
{\footnotesize
\begin{tabbing}
xxxx1\=xxxx\=xxxx\=xxxx\=xxxx\=xxxx\=\kill
\>1: HMM-SNN(N, k, P, T, signal): \\
\>2:\> data = Feature-Extraction(signal, N) \\
\>3:\> sub-patterns = Auto-Segmentation(data, P) \\
\>4:\> For each sample in sub-patterns: \\
\>5:\> \> spike-trains = Extract-Poisson-Spikes(pattern, T)  // e.g. 80 spike trains \\
\>6:\> \> Calculate output neuron status using Eq.~\ref{eq:exp_uk}\\
\>7:\> \> if (Training-Session): \\
\>8:\> \> \> Train SNNs using Eqs.~\ref{eq:NesslerSTDP_wts}-\ref{eq:pi_snn}\\
\>9:\> \> else\\
\>10:\> \> \> Select class with highest Pr using Eq.~\ref{eq:HMMrecognitionEqns} or Eq.~\ref{eq:HMMrecognitionEqnsSpeech} \\
\end{tabbing}}
\end{center}
\caption{Algorithm for hybrid model generation and classifying the sequential data proposed in this paper.}
\label{fig:modelAlgorithm}
\end{figure}

\subsection{Speech Signals}
This experiment extends the method to speech signal processing.
A speech signal can be characterized as a number of sequential frames with stationary characteristics
within the frame.
A speech signal $S=f_1 f_2 \ldots f_M$ has $M$ sequential frames.
In humans, the signal within the auditory nerve is the result of an ongoing
Fourier analysis performed by the cochlea of the inner ear.
That is, the frequencies' energy and formants carry useful information for the 
speech recognition problem.
In our experiments, we divided speech signals into 20 ms duration frames with 50 percent temporal
overlap and converted each frame to the frequency domain (Line~2 of Fig.~\ref{fig:modelAlgorithm}).

\begin{figure}
\begin{center}
{\footnotesize
\begin{tabbing}
xxxx1\=xxxx\=xxxx\=xxxx\=xxxx\=xxxx\=\kill
\>1:\> Initialize the data into $P$ equal-width segments (sub-patterns). \\
\>2:\> Repeat \\
\>3:\> \> sample=1 \\
\>4:\> \> For $p$=1 to $P-1$ \\
\>5:\> \>  \> While (distance(sample and centroid[$p$]) $\le$ distance(sample and centroid[$p+1$]) \\
\>6:\> \> \> \> sample++\\
\>7:\> \> \> Segment($p$)=Sample \\
\>8:\> \> \> Update centroids \\
\>9:\> Until Segment change $\le$ threshold. \\
\end{tabbing}}
\end{center}
\caption{Pseudocode for auto-segmentation preprocessing. $\mathrm{Segment}(p)$ specifies the last sample of each cluster.}
\label{fig:autoSegmentationPreProc}
\end{figure}

\paragraph{Auto segmentation preprocessing step}
The preprocessing groups the $M$ sequential frames into $P$ consecutive clusters.
Let 
$\mathbf{O} = [\mathbf{o}_1 \ldots \mathbf{o}_M]$
denote a sequence of observations that is a member of, say, class $C_1$.
The goal of the SNN is to classify $\mathbf{O}$ as a member of $C_1$ among the other 
possible classes.
For instance, $\mathbf{O}$ can be a speech stream with $M$ 20 ms duration frames, where
each $\mathbf{o}_m$ is a frame consisting
of $N$ features observed at a given 20 ms time step.
We shall call this a feature vector.
The $M$ feature vectors (frames) should map to the $P$ categories corresponding to the HMM states.

The initial problem is to cluster the $M$ vectors into the $P$ HMM categories.
For this purpose, a modified $k$-means algorithm was used.
The algorithm, given in Fig.~\ref{fig:autoSegmentationPreProc}, compares consecutive (adjacent) clusters to group the frames
into $P$ sequential data segments (Line~3 of Fig.~\ref{fig:modelAlgorithm}).
Each segment contains approximately similar feature vectors as judged by the clustering algorithm.

\begin{figure}
\begin{center}
  \includegraphics[viewport=1.5in 4.2in 7.5in 7in,clip=true,scale=0.78]{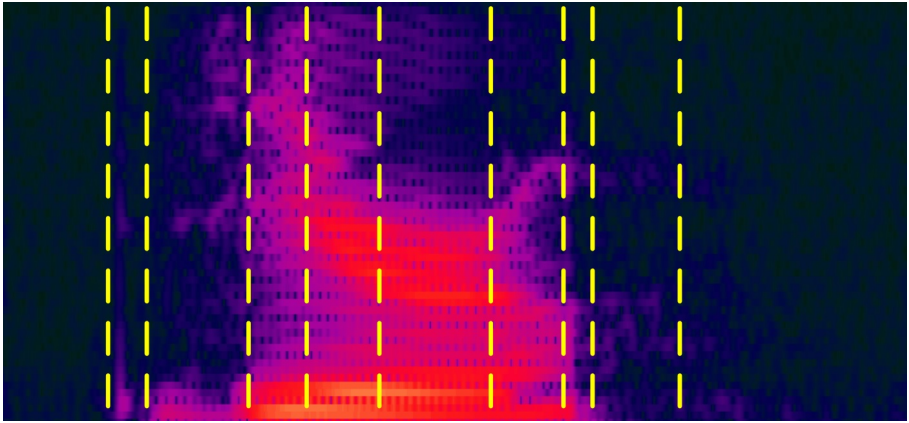}
\end{center}
\caption{Spectrogram for the spoken word ``zero.''
The horizontal and vertical axes represent the time and frequency, respectively.
Color represents power.
The vertical dashed lines represent boundaries between the $P=10$ segments subpatterns.
Each subpattern consists of many sequential samples.}
\label{fig:zeroSpectrogram}
\end{figure}

To illustrate, Fig.~\ref{fig:zeroSpectrogram} shows a spectrogram of the spoken word ``zero.''
This represents the power spectrum of frequencies in a signal as they vary 
with time.
The auto segmentation result for this signal is also shown in Fig.~\ref{fig:zeroSpectrogram} by the
vertical dashed lines.
The signal has been divided into $P=10$ segments containing a varying number of
speech frames.
A specific segment consists of similar frames, where the signal is approximately stationary,
and corresponds to one state of the HMM\@.

\paragraph{Converting a speech signal to a spike train}
The speech signals were sampled at 8 kHz.
Frame duration was taken to be 20 ms, which at an 8 kHz sampling rate
contains 160 sample values.
After converting to the frequency domain, this reduces to 80 sample values.
The magnitudes of the 80 frequency components were converted to rate parameters
for 80 Poisson spike trains (Line~5 of Fig.~\ref{fig:modelAlgorithm}). 
The simulation time for the spike trains was $T=20$ ms.

\paragraph{Classifying spoken words}
Two experiments were conducted, classifying spoken words into either two or four categories.
Data was selected from the Aurora dataset \cite{Pearce2000a}.
The data set contains spoken American English digits taken from male and female speakers
sampled at 8 kHz.
600 spoken digits belonging five categories ``zero'', 
``one,'' ``four,'' ``eight'', and ``nine'' were selected. 

For each word recognized, an HMM with $P=10$ states was used.
A separate SNN was associated with each HMM state.
All SNNs used 80 input units and 8 output units.
The network had $8\cdot81=648$ adaptive weights.
These dimensions are the same as the network used in the previous experiment.
Since the probability distribution functions of the states are independent of
each other, the $P$ speech segments can be trained in parallel.
The within-class variability for a single state is maintained by the $K=8$ output units which
approximate a Gaussian mixture model.
The input spike trains are obtained by the Poisson process based on the
frame's frequency amplitudes (80 feature values).
The model was trained for 100 iterations.
After training,
the synaptic weights reflect the importance of specific frequencies and 
the final bias weights show the output neurons' excitability in each state. 
The recognition phase in this experiment is more general than in Eq.~\ref{eq:HMMrecognitionEqns} 
such that each segment $S$ (1 through $P$=10), which is determined by a state, 
contains the number of samples. Thus,

\begin{subequations}
\label{eq:HMMrecognitionEqnsSpeech}
\begin{align}
\mathit{Pr}(\mathbf{O} | \boldsymbol\lambda) &= \mathit{Pr}(s_m =1) \prod_{p=2}^{P=10} a_{p-1,p} \cdot \mathit{Pr}( s_m = p ) \\ & 
\mathit{Pr}( s_m = p) = \prod_{l\in S(p)}a_{p,p}\cdot \prod_{t=1}^{T=20} \mathit{Pr}_{\mathrm{snn}_p}^{l}(t),
\end{align}
\end{subequations}
where $\mathit{Pr}_{\mathrm{snn}_p}^{l}(t)$ specifies the probability measure of sample $l$ of the 
sub-pattern corresponding to state $p$.

\begin{figure}
\begin{center}
  \includegraphics[viewport=0.0in 0.0in 7.5in 5in,clip=true,scale=0.5]{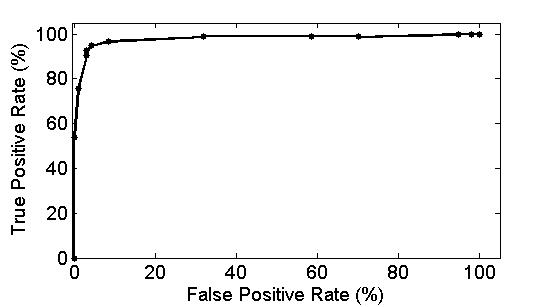}
\end{center}
\caption{ROC curve of the results for classifying the spoken words ``zero" versus ``one".}
\label{fig:ROC}
\end{figure}

Table~\ref{tab:binarySpeechResults} shows the binary classification performance
using different relative prior probabilities as bias parameters. Fig.~\ref{fig:ROC} illustrates the ROC curve of the results shown in Table~\ref{tab:binarySpeechResults}.
The accuracy rate above 95 percent shows initial success of the model.
Table~\ref{tab:fourClassSpeechResults} shows accuracy rates of the model in recognizing four spoken words.
An average performance above 85 percent accuracy was obtained.

\begin{table}
\centering
\caption{Results for classifying the spoken words ``zero'' versus ``one.''
The highest accuracy is 95.27 percent. ``One'' is arbitrarily chosen as the positive class.}
\label{tab:binarySpeechResults}       
\begin{tabular}{lrrl}
\hline\noalign{\smallskip}
$\frac{P(\textrm`0\textrm')}{P(\textrm`1\textrm')}$ & FP \% & TP \%  & Accuracy \% \\
\noalign{\smallskip}\hline\noalign{\smallskip}
0.9500 & 100.00 & 100.00 & 50.26\\
0.9600 & 97.87 & 100.00 & 52.91\\
0.9650 & 94.68 & 100.00 & 51.32\\
0.9830 & 70.21 &  98.95 & 64.55\\
0.9850 & 58.51 &  98.95 & 70.37\\
0.9900 & 31.94 &  98.95 & 83.60\\
0.9965 &  8.51 &  96.84 & 94.18\\
0.9980 &  4.26 &  94.74 & \textbf{95.27}\\
0.9990 &  3.19 &  92.63 & 94.71\\
1.0000 &  3.19 &  90.53 & 93.65\\
1.0030 &  1.06 &  75.79 & 87.30\\
1.0101 &  0.00 &  53.68 & 76.72\\
1.3333 &  0.00 &   0.00 & 49.00\\
\noalign{\smallskip}\hline
\end{tabular}
\end{table}

\paragraph{Summary of parameter choices}
The number of input units, $N=80$, was chosen because there were 80 frequency components
in the spectrogram at the sampling rate used.
The number of HMM states, $P=10$, was chosen as the smallest value to qualitatively
represent the variations in the acoustic structure of the spectrograms encountered.
The number of output units, $K=8$, was chosen to be the same as number of distributions considered for the GMM in previous study~\cite{Tavanaei2012a}.

\subsection{Discussion of results}
Previous work was conducted using a traditional support vector
data description and an HMM to classify spoken digits using wavelets
and frequency-based features \cite{Tavanaei2012a}.
Accuracy rates above 90 percent were achieved which is better than the 
results obtained in the present experiments.
However, that model does not have the biomorphic features that exist in
the present model. Additionally, online learning in the current method makes the model flexible to new data occurrences and is able to be updated efficiently.
Furthermore, using the SNNs which support communication via a series of the impulses 
instead of real numbers would be useful in VLSI implementation of the human brain functionality in sequential pattern recognition.

\begin{table}
\centering
\caption{Results for classifying the spoken words ``zero,'' ``four,''
``eight,'' and ``nine.''
The average classification accuracy is 85.57 percent.}
\label{tab:fourClassSpeechResults}       
\begin{tabular}{lrrl}
\hline\noalign{\smallskip}
Class & Accuracy \% \\
\noalign{\smallskip}\hline\noalign{\smallskip}
``zero'' & 81.91 \\
``four'' & 82.98 \\
``eight'' & 96.74 \\
``nine'' & 80.65 \\
\textbf{Average}  & \textbf{85.57}\\
\noalign{\smallskip}\hline
\end{tabular}
\end{table}

\section{Conclusion}
A novel hybrid learning model for sequential data classification was studied.
It consisted of an hidden Markov model combined with a spiking neural network
that approximated expectation maximization learning.
Although there have been other hybrid networks, to our knowledge this is the first
using an Snn\@.
The model was studied on synthesized spike-train data and also on spoken word data.
Although the studies are preliminary, they demonstrate proof-of-concept in the
sense that it provides a useful example of how a statistically based mechanism for
signal processing may be integrated with biologically motivated mechanisms, such as
neural networks.

Our approach derives from the described in \cite{Nessler2009a,Nessler2013a,Kappel2014a}.
The work in \cite{Nessler2009a,Nessler2013a} showed how to use STDP learning to approximate
expectation maximization.
A complete HMM was encoded in a recurrent spiking neural network in the work of \cite{Kappel2014a}.
Our approach seeks a middle ground where the recurrent neural network is unwrapped into 
a sequence of HMM states, which each state having an associated nonrecurrent network.
This leaves open the possibility of thinking about other ways to encode HMMs in brain
circuitry by refining a model that does not make such extreme assumptions.

Planned future work includes continuing to study the model's properties, both
theoretically and experimentally, improving and extending the model's range of performance,
and extending the learning capabilities of the model.
Conspicuously, the state transition probabilities of the present model are predetermined
and fixed.
Our most important goal is to learn the sequential data in an online fashion without the
segmentation preprocessing phase.
This would provide a basis to train the state transition probabilities.




\end{document}